\documentclass{article} %

\usepackage{hyperref}
\usepackage{url}
\RequirePackage{algorithm}
\RequirePackage{algorithmic}
\usepackage[english]{babel}
\usepackage[authoryear]{natbib}
\usepackage{microtype}
\usepackage{graphicx}
\usepackage{subcaption}
\usepackage{booktabs} %
\usepackage{wrapfig}
\usepackage{amsmath}
\usepackage{amssymb}
\usepackage{mathtools}
\usepackage{amsthm}
\usepackage{thmtools}
\usepackage{thm-restate}
\usepackage{xcolor}
\usepackage{listings}
\usepackage{caption}
\usepackage{wrapfig}
\usepackage{tcolorbox}
\usepackage{titletoc}
\usepackage{placeins}
\usepackage{multirow}
\theoremstyle{plain}

\theoremstyle{definition}

\theoremstyle{remark}

\usepackage[capitalize,noabbrev,nameinlink]{cleveref}
\crefname{assumption}{Assumption}{Assumptions}
\Crefname{assumption}{Assumption}{Assumptions}

\definecolor{codeblue}{rgb}{0.25,0.5,0.5}
\definecolor{codekw}{rgb}{0.85, 0.18, 0.50}

\definecolor{codegreen}{rgb}{0,0.6,0}
\definecolor{codegray}{rgb}{0.5,0.5,0.5}
\definecolor{codepurple}{rgb}{0.58,0,0.82}
\definecolor{backcolour}{rgb}{0.95,0.95,0.92}
\definecolor{codepurple}{rgb}{0.58,0,0.82}
\definecolor{papercolor}{HTML}{0668E1}
\definecolor{darkred}{rgb}{0.68,0.05,0.0}
\definecolor{tab_blue}{RGB}{230,245,255} %
\definecolor{tab_purple}{RGB}{245,230,255} %
\definecolor{mygreen}{RGB}{66,150,83}

\usepackage{ssArxiv}

\hypersetup{
    colorlinks,
    linkcolor={mygreen},
    citecolor={darkblue},
    urlcolor={codepurple}
}
\newcommand{\samecolorfootnote}[1]{\textsuperscript{\textcolor{darkred}{#1}}}
\renewcommand{\thefootnote}{\textcolor{darkred}{*}}

\renewcommand{\thefootnote}{\arabic{footnote}}

\title{When Less is Enough: \\  Efficient Inference via Collaborative Reasoning}

\author{
\begin{tabular}{c@{\hspace{1.2cm}}c@{\hspace{1.2cm}}c}
{Yilei Chen
\samecolorfootnote{*}
} &
{Sharut Gupta
\samecolorfootnote{*}
} &
{Yannis Paschalidis} \\
Boston University &
MIT CSAIL &
Boston University  \\
\texttt{ylchen9@bu.edu} &
\texttt{sharut@mit.edu} &
\texttt{yannisp@bu.edu} \\
\\[3.0ex]
\multicolumn{3}{c}{%
\begin{tabular}{c@{\hspace{2.2cm}}c}
\textbf{Ayush Sekhari\samecolorfootnote{$\dagger$}} &
\textbf{Aldo Pacchiano\samecolorfootnote{$\dagger$}} \\
Biohub &
Boston University \\
\texttt{ayush@sekhari.com} & 
\texttt{pacchian@bu.edu}
\end{tabular}
}
\end{tabular}
}

\usepackage{enumitem}
\usepackage[suppress]{color-edits}  
\addauthor{as}{purple} 

\begin{document} 
\maketitle 

\begin{abstract}
 In this work, we introduce DUET  (\textbf{Du}al-model \textbf{E}fficient \textbf{T}wo-stage inference), a collaborative inference framework in which a capable model and a lightweight model work together to solve a task. Relying on a single large model to perform end-to-end reasoning and prediction often incurs substantial inference cost. In contrast, DUET decomposes inference into two stages: the capable model produces a reasoning signal, and the lightweight model interprets this signal to generate the final answer, allowing reasoning-intensive computation to be handled by the capable model while non–reasoning-intensive components are delegated to the lightweight model without sacrificing task performance. To achieve this objective, we propose a length-penalized joint training objective that encourages the capable model to transmit only the information that is sufficient for the lightweight model to solve the task. As a result, DUET maintains strong reasoning performance with substantially lower inference cost than end-to-end inference using a large model alone, saving up to 60\% of the large model’s output tokens on challenging reasoning benchmarks, including AIME and GPQA. We open-source the code at: \url{https://github.com/fairytale9/llm_bottleneck}.
\end{abstract}

\begingroup
\renewcommand{\thefootnote}{}
\footnotetext{$*$ denotes equal first-author contribution. $\dagger$ denotes equal advisory role.}
\endgroup

\section{Introduction}

\noindent

\noindent 
To solve complex tasks such as mathematical or scientific reasoning, models increasingly rely on long reasoning chains in which the generation of long and extensive intermediate reasoning traces is correlated with accurate prediction~\citep{wei2022chain,muennighoff2025s1,guha2025openthoughts,wang2022self,nye2021show,kaplan2020scaling,hoffmann2022training,wang2022self,wei2022chain,nye2021show,sardana2023beyond,chowdhery2023palm,snell2024scaling,wu2024inference}. While effective, this approach to inference can prove computationally prohibitive. For example, on AIME 2024, reasoning models produce responses 5-10x longer than equivalently-sized instruct models, even on questions that both can solve correctly~\citep{fan2025price}. Similarly, DeepSeek-R1 generates 31x more intermediate tokens than Llama 3.1 8B while running 54x slower, requiring 16 H100 GPUs just to serve~\citep{garvey2025deepseekr1inference}.

Several directions have been explored to improve this compute-performance tradeoff in LLMs. Methods such as model distillation~\citep{hinton2015distilling} and supervised fine-tuning~\citep{muennighoff2025s1,ye2025limoreasoning,guha2025openthoughts} reduce model size and thus the inference cost, but often fail to match the performance of larger models~\citep{liu2025dler}. Other approaches including pruning~\citep{nayab2024concise} and constrained chain-of-thought generation~\citep{munkhbat2025self,fatemi2025concise,gao2025concise,song2025walk} aim to directly reduce the generated token count but risk discarding information critical for correctness~\citep{lajewska2025understanding}.\looseness=-1

\begin{figure*}[!htb]
    \centering
    \begin{subfigure}[t]{1.\linewidth}
        \centering
        \includegraphics[width=\linewidth]{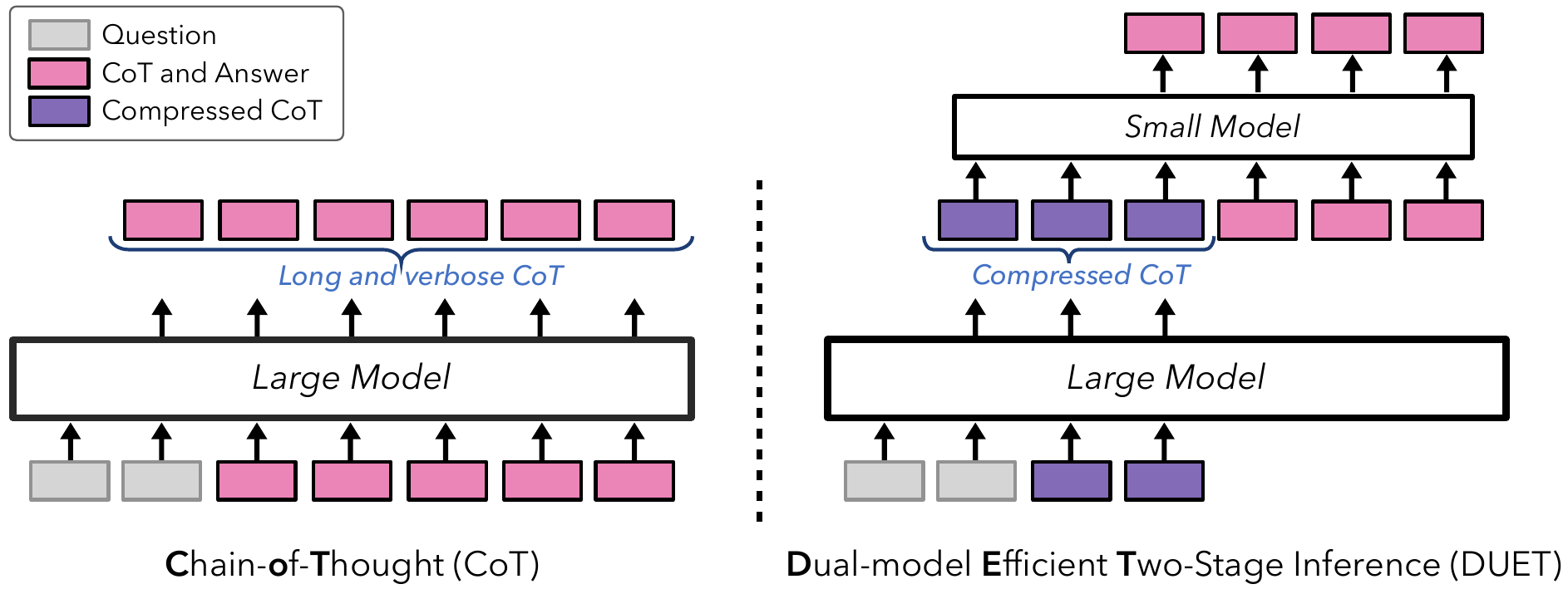}
    \end{subfigure}
    \hfill
    \caption{\textbf{High-level overview of the DUET framework.} Standard CoT relies on a large model to generate long reasoning traces, substantially increasing inference cost. DUET instead has the capable model produce a high-level reasoning signal that a lightweight model uses to generate the final answer, reducing generation length (and cost) without sacrificing accuracy.
    }
    \label{fig:teaser}
\end{figure*}

\begin{wrapfigure}{r}{0.5\textwidth} %
    \centering
    \includegraphics[width=0.5\textwidth]{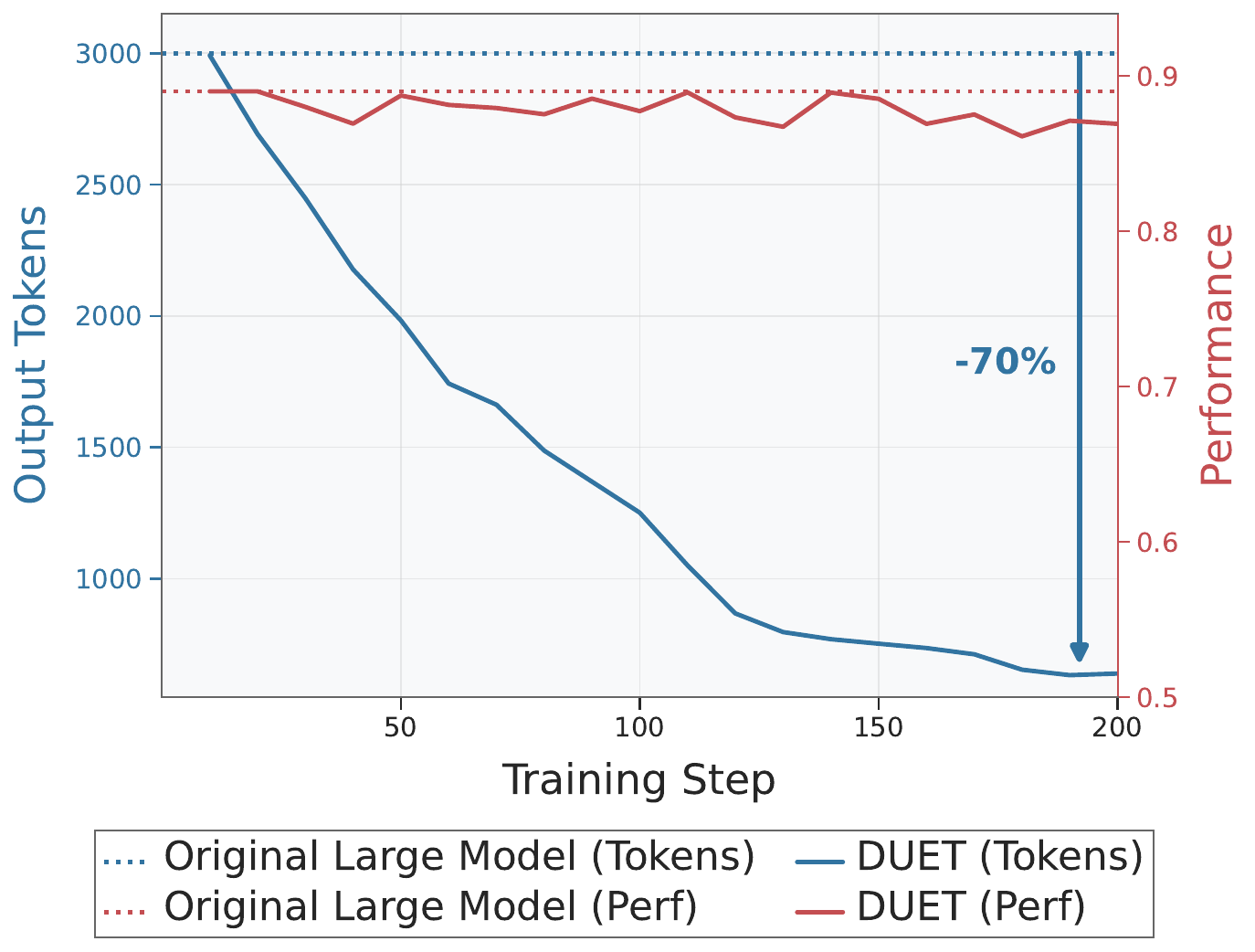}
    \caption{The number of output tokens generated by large model M vs performance over DUET iterations. DUET achieves a reduction in output length by up to 70\% with only 200 steps of training. At the same time, performance is comparable to the original large model.} 
    \label{fig:intro_duet}
\end{wrapfigure} 

Taken together, these limitations suggest that the inefficiency is not merely a byproduct of scale, but reflects a more fundamental architectural constraint.
In particular, these models assign equal computational weight to all components of a task---whether a complex logical deduction or trivial syntactic formatting---and entangle reasoning with response generation in a single token stream. This is in contrast to 
how humans and other organisms allocate cognitive resources, and violates the basic principle of \emph{cognitive economy}~\citep{lieder2020resource,attwell2001energy,barlow1961possible} that establishes that biological organisms allocate metabolic resources in a manner proportional to task difficulty. In fact, this separation not only reduces total effort but also produces compressed, transferable representations. Unfortunately, the current reasoning models forego these benefits of distributed labor. \looseness=-1

In this work, we move beyond this monolithic design and propose a hierarchical division-of-labor principle, in which different parts of the inference process are handled by different models according to their computational demands. Consider, for example, a complex mathematical problem. Experienced mathematicians often begin by identifying a high-level proof strategy; only after this plan is in place do they turn to the more laborious work of filling in details, checking intermediate claims, and carrying out the necessary calculations. Taking inspiration from this example, we introduce DUET (\textbf{Du}al-model \textbf{E}fficient \textbf{T}wo-stage inference, ~\Cref{fig:teaser}), a collaborative post-training framework that decomposes inference into two stages: (1) reasoning compression, where a large model $M_\theta$ outputs a compressed reasoning signal $z$, and (2) response generation, where a lightweight model $m_\phi$ conditions on $z$ to produce the final answer.

In practice, this is achieved by simultaneously training both models using a length-penalized objective that explicitly minimizes the number of tokens generated while maintaining accuracy. The objective combines (1) a marginal-utility reward that is positive only when the small model is likely to fail, thereby incentivizing use of the large model precisely when it improves outcomes, and (2) an adaptive length-penalty schedule that penalizes total token generation. Empirically, on AIME and GPQA~\citep{rein2024gpqa}, \textsc{DUET} matches large-model test performance while reducing its token usage by up to roughly $60\%$.

\noindent Our key contributions are: 
\begin{enumerate}
    [label=$\bullet$]  
  \item We introduce DUET, a dual-model, two-stage inference paradigm that decouples capability-intensive reasoning from inexpensive response generation. %
  \item We propose a length-penalized joint training objective with alternating optimization. We augment this with marginal-utility rewards and an adaptive length-penalty schedule for stability and budget control.
  \item We demonstrate that DUET substantially reduces expensive model tokens, up to $60\%$ on challenging benchmarks including AIME and GPQA, while preserving the baseline performance.
\end{enumerate}

\section{Related Work}

We present only a subset of the related works here and provide additional discussion in~\Cref{app:related_work}.

\noindent \textbf{Model collaboration.} DUET belongs to a broader class of frameworks that leverage collaboration among multiple models. One prominent line of research is speculative decoding, which accelerates generation by allowing a small model to first produce draft tokens that are subsequently verified or refined by a larger model \citep{chen2023accelerating, chen2024cascade}. Another widely studied paradigm is model routing, where a learned router assigns queries to the small or large model based on the predicted query difficulty and the desired quality level \citep{ding2024hybrid, mei2025omnirouter, behera2025towards}. Similarly, model cascading \citep{kolawole2024agreement} builds a cascade of models of different sizes to dynamically assign inputs for efficient inference.

\noindent \textbf{Concise reasoning.}  A growing body of work investigates efficient reasoning. One line of research explicitly encourages conciseness through fine-tuning or optimization objectives. \citet{munkhbat2025self} showed that models can produce concise reasoning traces by fine-tuning on self-generated examples. \citet{fatemi2025concise} observed that standard reinforcement learning objectives tend to favor overly long responses and introduce a secondary RL fine-tuning phase explicitly rewarding brevity. \citet{song2025walk} proposed a two-stage RL algorithm which first trains the model for more steps, and subsequently fine-tunes it for conciseness in the second stage. \citet{gao2025concise} considered a length-constrained optimization problem closely related to ours, however, their approach considers a single model and enforces a hard constraint on the token budget during generation, whereas DUET adopts a dual-model framework that allows for more flexible reasoning compression. \asedit{\citet{arora2025training} also train reasoning models to reduce inference-time compute using a reinforcement-learning objective that incentivizes shorter chains of thought while maintaining accuracy, yielding a family of models with different efficiency–accuracy trade-offs. Similar to \citet{gao2025concise}, this approach operates in a single-model setting, whereas DUET uses a dual-model decomposition in which a large model communicates a compressed reasoning signal to a lightweight model.}

\noindent \textbf{Information bottleneck.} The way DUET decomposes answer generation into an initial reasoning trace produced by a large model and a subsequent, potentially longer, generation by a smaller model is closely related to the information bottleneck principle. The information bottleneck (IB) principle was originally introduced by \citet{tishby2000information} and later extended to deep learning models \citep{tishby2015deep}. More recently, several works have used information-theoretic perspectives to study and improve reasoning in large language models. \citet{lei2025revisiting}, for example, view reasoning as an information bottleneck and use this perspective as a regularizer to encourage reasoning trajectories to remain informative about the correct answer. \citet{ton2024understanding} analyze chain-of-thought reasoning through an information-theoretic lens by quantifying the information gain associated with each reasoning step. Similarly, \citet{wang2025learning} propose an information-theoretic reinforcement fine-tuning framework that encourages accurate reasoning with fewer generated tokens.

However, these approaches focus on single-model settings and treat the information bottleneck primarily as a regularization or analytical principle. By contrast, DUET uses the bottleneck principle structurally: the large model produces a compressed reasoning signal that must retain the information needed by a lightweight model to generate the final answer. In this sense, DUET instantiates an information-bottleneck-like decomposition of reasoning across two collaborating models, rather than using the bottleneck solely as a regularizer within a single model.

\begin{figure*}[htb]
    \centering
    \includegraphics[width=0.9\linewidth]{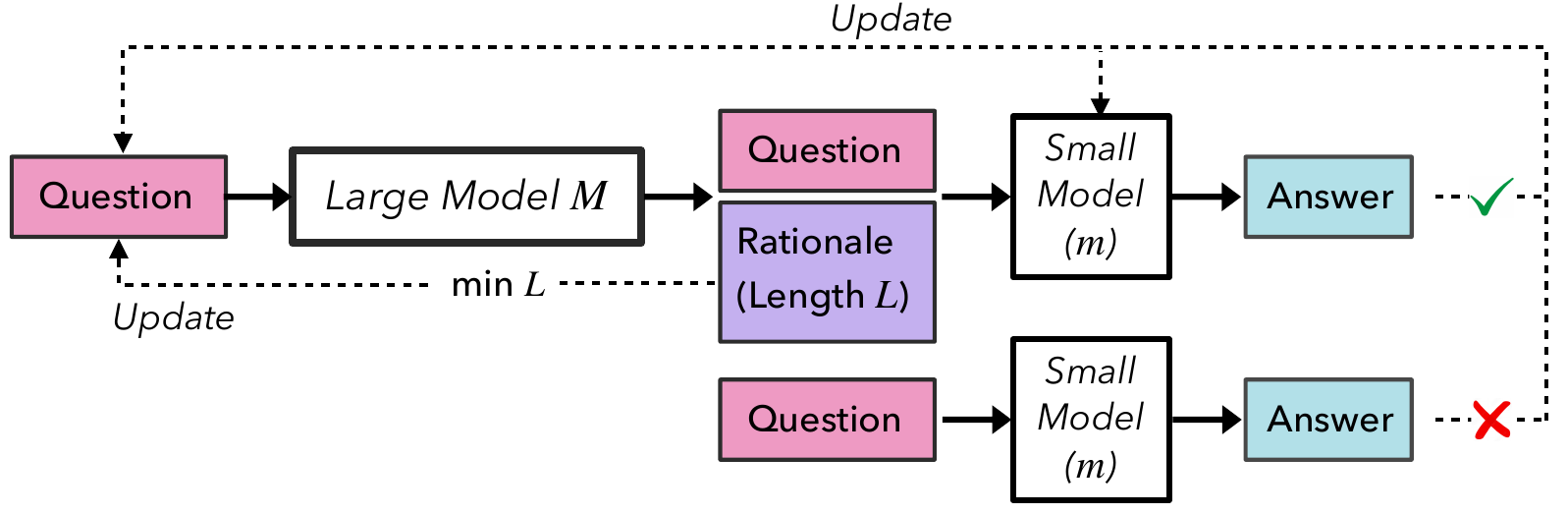}
    \caption{ \em \textit{For each input, the capable model generates a reasoning signal and the lightweight model produces an answer conditioned on (input + signal). In parallel, the lightweight model also produces a baseline answer from the input alone. The baseline score provides a reference point, and the gain achieved by adding the reasoning signal is used to update both models. The capable model is trained with an additional length penalty to keep its reasoning traces short.}\looseness=-1}
    \label{fig:pipeline}
\end{figure*}

\section{\underline{DU}al-Model \underline{E}fficient \underline{T}wo-stage Inference (DUET)} 

In this work, we propose DUET, a framework for efficient reasoning through collaboration between a large and a small model. DUET consists of two pretrained models: a capable (large) model $M_\theta$ and a lightweight (small) model $m_\phi$. When the context is clear, we omit parameter subscripts for simplicity.

\noindent Given an input prompt $x$, the large model first produces an intermediate representation 
\begin{align*}
    z\sim M(\cdot|x),
\end{align*}
which serves as a communication medium between the two models. Conditioned on both the input $x$ and the intermediate representation $z$, the small model then generates the final response,
\begin{align*}
    y_{Mm}\sim m(\cdot|x,z).
\end{align*}

The most direct application of the DUET framework is efficient inference. In this setting, $z$ corresponds to the reasoning process generated by the large model, while the small model conditions on this reasoning to produce the final answer. This decouples expensive reasoning from response generation, allowing inference cost to be significantly reduced.

Individually, the large model $M$ is capable of solving complex tasks but typically incurs high inference cost due to model complexity (large number of parameters) and the length of the reasoning traces. In contrast, the small model $m$ is computationally efficient but lacks sufficient capacity to reliably solve the given tasks on its own. DUET is a way to combine their strengths: the large model learns to handle capability-intensive reasoning, while the small model performs efficient response generation, thereby achieving a favorable trade-off between performance and inference efficiency. \ascomment{Some examples of the outputs generated by M (the large model) and m (the small model) can be found in Appendix~\ref{app:examples}.}

\subsection{Algorithm} 
\label{sec:alg}

DUET proposes the following constrained optimization objective:
\begin{equation}\label{eq:constrained_duet}
    \max_{M_\theta, m_\phi}  \mathbb{E}_{x, z\sim M_\theta(\cdot |x) }\left[\mathbb{E}_{y_{Mm}\sim m_\phi(\cdot|z)}[R(x,y_{Mm})] \right] \text{ s.t. } \mathbb{E}_{x, z\sim M_\theta(\cdot |x) }[ L(z) ] \leq B, 
\end{equation}
for some bound $B$ on the expected length of the generated sequence of intermediate tokens. \ascomment{here!} Intuitively, the objective in \eqref{eq:constrained_duet} corresponds to finding parameters for $M_\theta$ and $m_\phi$, such that they can maximize the expected reward, while keeping the expected length of intermediate representation (tokens) to be smaller than $B$. The expected length of $z$ directly correlates with inference-time computational cost, making it a natural objective for efficient reasoning. While the reward maximization objective ensures the total reward of the combined pipeline (i.e.~$x$ to $z$ to $y_{Mm}$) remains high.

In practice, however, we optimize an approximate (Lagrangian) version of \Cref{eq:constrained_duet}: 
\begin{align}
\max_{M_\theta} \max_{m_\phi} \mathbb{E}_{x\sim \mathcal{D}} \left[ \mathbb{E}_{z\sim M_{\theta}(\cdot|x)}[ \mathbb{E}_{y\sim m_\phi(\cdot|z)}[R(x,y)] - \lambda L(z) ] \right]. 
\label{eq:obj}
\end{align} 
where $\lambda >0$ corresponds to a regularization parameter. We optimize this objective via alternating optimization over the large model $M_\theta$ and the small model $m_\phi$. The pseudo-code is provided in~\Cref{app:algorithm}. During training step $t$, we collect a batch of samples generated by the two model pipeline,
i.e.~$\{x, z_M, y_{Mm}\}$, where $z_M \sim M_\theta(\cdot \mid x)$ and $y_{Mm} \sim m_\phi(\cdot \mid x, z_M)$.
We then optimize $M$ and $m$ alternatingly via the following objectives: 
\begin{equation}
\begin{aligned}
M:\quad
&\max_{M_\theta}\mathbb{E}_{x, z_M, y_{Mm}}
\big[ R(x, y_{Mm}) - \lambda \frac{L(z_M)}{B} \big], \\
m:\quad
&\max_{m_\phi}\mathbb{E}_{x, z_M, y_{Mm}}
\big[ R(x, y_{Mm}) \big].
\end{aligned}
\label{eq:model_obj}
\end{equation}
where $B$ can be thought of as the hyper-parameter controlling the target expected length bound from \Cref{eq:constrained_duet}. To further improve training stability and to avoid degenerate solutions, we introduce two practical techniques on top of the objective in \Cref{eq:model_obj}. In~\Cref{subsec:ablation}, we report ablation results showing how these two design choices affect the performance of DUET.

\paragraph{Marginal Utility.} 
In our experiments, we observed that directly optimizing $M$ with the reward $R(x, y_{Mm}) - \lambda \frac{L(z_M)}{B}$ 
can lead to poor convergence. This issue was particularly pronounced on tasks for which the small model $m$ already achieves relatively strong baseline performance. In such settings, optimizing \Cref{eq:obj} can drive the large model $M$ to emit extremely short reasoning traces, effectively offloading most of the generation to the small model. Intuitively, because the objective is non-convex, the relative scaling of the task reward $R$ and the compression cost $L$ may favor a degenerate solution in which the large model minimizes communication by setting $L(z_M)$ close to zero, even when doing so provides little useful guidance. 

To mitigate this issue, we replace the reward term $R(x, y_{Mm})$ with a marginal utility reward $R(x, y_{Mm}) - R(x, y_m)$
where $y_m \sim m(\cdot \mid x)$ denotes the output of the small model when run without access to $z_M$. This yields the modified training objective $\left(R(x, y_{Mm}) - R(x, y_m)\right) - \lambda \frac{L(z_M)}{B}$. This marginal reward directly measures the contribution of the large model’s reasoning by quantifying how much access to $z_M$ improves the small model’s performance over its standalone baseline.

\paragraph{Length Penalty Schedule.} 
Although \Cref{eq:obj} is a reasonable relaxation of the constrained problem in \Cref{eq:constrained_duet}, a more faithful approach is to adapt the Lagrange multiplier online according to the degree of constraint violation. Concretely, we use the dual-ascent update
\[
\lambda_t = \Bigl[\lambda_{t-1} + \eta\left(\frac{\mathbb{E}[L(z_M)]}{B} - 1\right)\Bigr]_+,
\]
where $\eta>0$ is a stepsize parameter and $[\cdot]_+$ denotes projection onto the nonnegative reals. This is the standard primal-dual update for inequality-constrained optimization, specialized here to the budget constraint $\mathbb{E}[L(z_M)] \le B$ (see for example~\cite{boyd2004convex}).

The adaptive length penalty introduces a curriculum, prevents early collapse of the large model that can occur with a fixed $\lambda$, and eliminates the need for extensive hyperparameter tuning over penalty strengths. In practice, this schedule leads to more stable training and smoother convergence.

\section{Experiments}
In this section, we present the experimental results of DUET, with the goal of answering the following questions:
\begin{itemize}  
\item Is DUET an effective framework for efficient inference?
\item How sensitive is DUET to the choice of training dataset?
\item How do different designs within DUET affect its performance?
\end{itemize}
\subsection{Experimental Setting}
\noindent \textbf{Models.} We adopt Qwen3-4B (thinking mode enabled) as the capable (large) model and Qwen3-0.6B  as the lightweight (small) model \citep{yang2025qwen3}. Given our limited computing resources of only 4 H100 GPUs, Qwen3-4B is a suitable choice for the large model in DUET, as it offers a strong balance between reasoning capability and training efficiency: it possesses competitive original performance on challenging reasoning benchmarks while remaining feasible to train efficiently on four H100 GPUs. In contrast, Qwen3-0.6B is significantly more lightweight and cost-effective, making it a natural fit for the small model in DUET framework.

\noindent \textbf{Training and Evaluation.} Both models are trained using the standard GRPO algorithm \citep{shao2024deepseekmath}. Training is conducted on one of three datasets: MATH-LightEval (train split), DeepScaleR, or DAPO-Math-17k \citep{hendrycksmath2021, deepscaler2025, yu2025dapoopensourcellmreinforcement}. We evaluate the models on five challenging reasoning benchmarks: MATH500, AMC23, AIME2024, AIME2025, and GPQA-Diamond \citep{hendrycksmath2021, rein2024gpqa}. 

\noindent \textbf{Metrics.} We report results using three evaluation metrics: \textit{Accuracy, Token length\footnote{We report the token length of the large model's responses. The small model's average response length is below 1,000 tokens, which corresponds to roughly fewer than 100 tokens for the large model at equivalent inference cost.}, and Intelligence-per-Token} (IPT) \citep{liu2025dler}. Accuracy measures the task performance on each benchmark. Token length corresponds to the total number of output tokens generated by the large model. Finally, \textbf{IPT is defined as the ratio of accuracy to token length}, capturing the trade-off between the task performance and reasoning efforts. Note that a higher IPT score does not necessarily imply better overall performance; rather, IPT serves as an auxiliary efficiency measure and should be interpreted jointly with accuracy and token length.

\noindent \textbf{Baselines.}
We compare DUET against the large model's original performance, in which the large model conducts end-to-end reasoning and answering alone. In addition, we include a naive chain-of-thought truncation baseline. In this setting, a fixed reasoning budget is pre-specified, and the large model is prompted to generate reasoning as usual; once the budget is exhausted, the reasoning process is manually truncated. The truncated trace, together with the original input, is then passed to the large model itself, which generates the final answer. %

Beyond the naive baselines, we include four additional baselines that explicitly target concise reasoning, spanning both prompt-based and training-based approaches: Prompt-based, Single-model, ThinkPrune \citep{hou2025thinkprune}, and L-GRPO \citep{song2025walk}. Detailed descriptions and implementation specifics are provided in Appendix~\ref{app:baseline}.

\noindent Other detailed training configurations, including batch sizes, learning rates, and other hyperparameters, are provided in~\Cref{app:exp_setup}.

\ascomment{Need to do main figure 1 again!} 
\begin{figure*}[htb]
    \centering
\includegraphics[width=\linewidth]{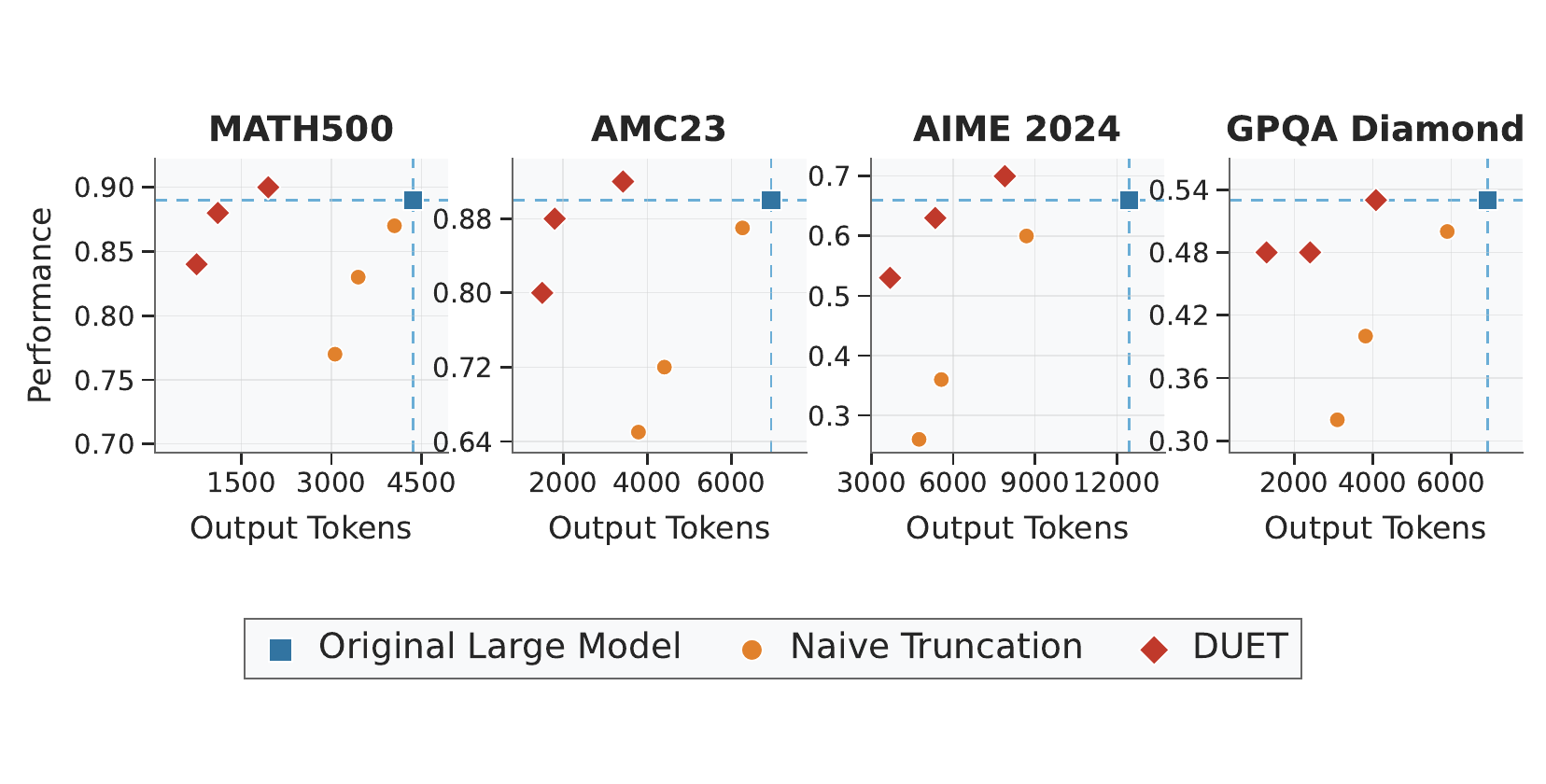}
    \caption{ \em \textit{We compare DUET with the original large model and naive truncation on MATH500, AMC23, AIME2024, and GPQA Diamond benchmarks. The x-axis shows the total number of output tokens generated by the large model, and the y-axis shows task performance. Dashed vertical and horizontal lines indicate the output tokens and performance of the original large model, respectively. Each DUET point corresponds to a checkpoint of different training steps (50, 100, and 150). Points for naive truncation correspond to fixed reasoning budgets of 3072, 4096, and 8192 tokens. DUET achieves performance comparable to or better than the large-model baseline while reducing the large-model output tokens by up to 75\%, demonstrating a favorable efficiency–performance trade-off.}} 
    \label{fig:main}
\end{figure*}

\subsection{Main Results And Ablations}  

We summarize our main results in~\Cref{fig:main}. Across all evaluated benchmarks, DUET consistently achieves performance comparable to, and in some cases exceeding, that of the original large model, while substantially reducing inference cost quantified by the output token length\footnote{Output token length refers to tokens generated by the large model. The small model adds negligible cost due to its smaller size and shorter outputs, so large-model decoding dominates inference cost.}. On average, DUET reduces the total number of output tokens generated by the large model by approximately $70\%$, demonstrating a favorable performance--efficiency trade-off enabled by DUET and joint training. 

The large model alone is capable of solving these reasoning tasks, but its performance typically relies on generating long reasoning traces. Naive truncation reduces inference cost by imposing a fixed reasoning budget, but it discards reasoning tokens indiscriminately, regardless of their importance. As a result, truncation often removes critical intermediate information, leading to noticeable performance degradation, especially on more challenging benchmarks.

In contrast, DUET explicitly trains the capable model to compress its reasoning into a concise and informative signal that is tailored to the lightweight model. Through joint training with a length-penalized objective, the capable model learns to preserve the most task-relevant information while eliminating unnecessary verbosity. This enables DUET to maintain strong reasoning performance while consuming much less tokens, highlighting the effectiveness and advantage of our DUET framework. \ascomment{We provide some example generations in \Cref{app:examples}.}

Unlike approaches that aim to improve a model’s capability, DUET is designed to preserve the existing capability of a strong model while reducing the amount of reasoning efforts required to realize that capability via collaboration with the small model. To study how sensitive DUET is to the choice of training data, we jointly train the dual-model system on three different datasets: MATH-LightEval, DeepScaleR, and DAPO-Math-17k. The results are summarized in~\Cref{tb:dataset}.

\begin{table}[htb]
\centering
\caption{ \em \textit{Performance of DUET compared with different baselines across benchmarks. Each cell reports Intelligence-per-Token (IPT), followed by (Accuracy, Large-model output tokens). IPT is measured as intelligence per 1000 tokens.}}  
\label{tb:baseline}
\vspace{0.5em}
\begin{tabular}{lcccc}
\toprule
\multirow{2}{*}{Method}
& \multicolumn{4}{c}{Benchmark} \\
\cmidrule(lr){2-5}
& AMC 23 & AIME 2024 & GPQA Diamond & MATH 500 \\
\midrule
\textbf{Original model}
& \textbf{0.130} {\scriptsize (0.90, 6948)}
& \textbf{0.053} {\scriptsize (0.66, 12458)}
& \textbf{0.077} {\scriptsize (0.53, 6928)}
& \textbf{0.204} {\scriptsize (0.89, 4358)} \\

\textbf{DUET}
& \textbf{0.489} {\scriptsize (0.88, 1800)}
& \textbf{0.118} {\scriptsize (0.63, 5337)}
& \textbf{0.200} {\scriptsize (0.48, 2402)}
& \textbf{0.796} {\scriptsize (0.88, 1106)} \\

\textbf{Prompt-based}
& \textbf{0.145} {\scriptsize (0.90, 6209)}
& \textbf{0.061} {\scriptsize (0.67, 11025)}
& \textbf{0.121} {\scriptsize (0.52, 4303)}
& \textbf{0.258} {\scriptsize (0.89, 3444)} \\

\textbf{Single model}
& \textbf{0.390} {\scriptsize (0.82, 2103)}
& \textbf{0.086} {\scriptsize (0.50, 5786)}
& \textbf{0.229} {\scriptsize (0.45, 1969)}
& \textbf{0.764} {\scriptsize (0.87, 1139)} \\

\textbf{ThinkPrune}
& \textbf{0.162} {\scriptsize (0.87, 5353)}
& \textbf{0.053} {\scriptsize (0.53, 10007)}
& \textbf{0.079} {\scriptsize (0.44, 5534)}
& \textbf{0.273} {\scriptsize (0.87, 3187)} \\

\textbf{L-GRPO}
& \textbf{0.162} {\scriptsize (0.92, 5668)}
& \textbf{0.067} {\scriptsize (0.66, 9835)}
& \textbf{0.101} {\scriptsize (0.54, 5332)}
& \textbf{0.260} {\scriptsize (0.89, 3429)} \\
\bottomrule
\end{tabular}
\end{table}

\subsubsection{Comparison Relative to Baselines}

We compare DUET with several prompt-based and training-based baselines designed to encourage concise reasoning. As shown in Table~\ref{tb:baseline}, DUET consistently outperforms all baselines across benchmarks, maintaining the high accuracy of the original model while using significantly fewer large-model tokens. For example, on AMC23, DUET improves IPT from 0.130 to 0.489 and reduces token usage from 6948 to 1800 while maintaining the original performance, with similar gains observed on AIME24, GPQA Diamond, and MATH500. Overall, DUET achieves the best efficiency–performance trade-off, reducing large-model tokens by up to 60–70\% while maintaining or even improving accuracy.

In contrast, the baseline approaches provide limited gains: prompt-based methods yield only modest improvements on the efficiency while single-model and ThinkPrune may degrade performance, indicating that naive compression either discards essential reasoning or fails to achieve truly concise reasoning. Beyond performance, DUET offers additional advantages. Its dual-model design decouples reasoning from final task completion, enabling the large model to produce compact, task-relevant signals while allowing the small model to be flexibly prompted or fine-tuned for different objectives—something that is difficult to achieve with a single concise model. Moreover, DUET is well-suited for practical deployment: a large model can reside on a remote server and provide a single reasoning signal, after which a lightweight local model performs downstream tasks on-device, enabling more efficient, privacy-preserving, and secure inference.

\begin{table*}[htb]
\centering
\caption{ \em \textit{Performance of DUET trained on different datasets across benchmarks. Each cell reports Intelligence-per-Token (IPT), followed by (Accuracy, Large-model output tokens). IPT is measured as intelligence per 1000 tokens.
\textbf{Top row:} original large model (end-to-end) performance.
\textbf{Bottom rows:} DUET trained on different datasets.}}
\label{tb:dataset}
\vspace{0.5em}
\begin{tabular}{lccccc}
\toprule
\multirow{3}{*}{Method / Training Data}
& \multicolumn{5}{c}{Benchmark} \\
\cmidrule(lr){2-6}
& MATH & AMC & AIME & AIME & GPQA \\
& 500  & 23  & 2024 & 2025 & Diamond \\
\midrule
\textbf{Original large model}
& \textbf{0.20} {\scriptsize (0.89, 4358)}
& \textbf{0.13} {\scriptsize (0.90, 6948)}
& \textbf{0.05} {\scriptsize (0.66, 12458)}
& \textbf{0.04} {\scriptsize (0.53, 12625)}
& \textbf{0.08} {\scriptsize (0.53, 6928)} \\
\midrule
\multicolumn{6}{l}{\textbf{DUET trained on:}} \\
\quad MATH-LightEval
& \textbf{0.81} {\scriptsize (0.90, 1109)}
& \textbf{0.45} {\scriptsize (0.85, 1878)}
& \textbf{0.13} {\scriptsize (0.63, 5013)}
& \textbf{0.07} {\scriptsize (0.47, 6418)}
& \textbf{0.19} {\scriptsize (0.52, 2689)} \\
\quad DeepScaleR
& \textbf{0.80} {\scriptsize (0.88, 1106)}
& \textbf{0.49} {\scriptsize (0.88, 1800)}
& \textbf{0.12} {\scriptsize (0.63, 5337)}
& \textbf{0.08} {\scriptsize (0.47, 5887)}
& \textbf{0.20} {\scriptsize (0.47, 2402)} \\
\quad DAPO-MATH-17k
& \textbf{0.69} {\scriptsize (0.88, 1271)}
& \textbf{0.33} {\scriptsize (0.85, 2544)}
& \textbf{0.10} {\scriptsize (0.57, 5898)}
& \textbf{0.05} {\scriptsize (0.43, 7819)}
& \textbf{0.16} {\scriptsize (0.52, 3190)} \\
\bottomrule
\end{tabular}
\end{table*}

\subsubsection{Generalization Across Difficulty Levels} 
A notable observation from~\Cref{tb:dataset} is that DUET exhibits strong generalization across substantial difficulty gaps between training and evaluation problems. When trained on MATH-LightEval, DUET successfully performs efficient reasoning while maintaining performance comparable to the original large model on significantly harder benchmarks such as AIME2024 and AIME2025, reducing large-model token usage by $60\%$ and $49\%$, respectively.

This behavior is fundamentally different from training for capability. Due to the large difficulty gap between MATH-LightEval and AIME-level problems, it is typically infeasible to train a model to solve AIME problems by learning directly from MATH-LightEval alone. In contrast, DUET does not attempt to acquire new problem-solving skills; instead, it learns to compress and distill the reasoning required to express the model’s existing intelligence.
These results provide strong empirical evidence that DUET is not training for improved capability, but rather for compressed intelligence—enabling the model to achieve similar accuracy with substantially fewer reasoning tokens.

\subsubsection{Sensitivity to Training Data}
 As shown in~\Cref{tb:dataset}, DUET trained on all three datasets consistently achieves substantially higher inference efficiency while maintaining task performance comparable to the original large model, as evidenced by markedly higher IPT (Information Per Token) and accuracy. 

Notably, the large model $M$ has original accuracies of $89\%$, $61\%$ and $77\%$ on Math-LightEval, DAPO-Math-17k and DeepScaleR, respectively. These results indicate that  DUET is effective across diverse training datasets as long as the large model exhibits sufficient baseline capability on the training data. Since DUET is designed to compress existing intelligence rather than to improve a model’s underlying capability, a natural question arises: 
\begin{center}
\emph{Can DUET be used to improve performance on problems where the large model itself struggles?} 
\end{center}

\begin{figure}[!htb]
    \centering
    \begin{minipage}{0.490\textwidth}
    \centering
    \includegraphics[width=\textwidth]{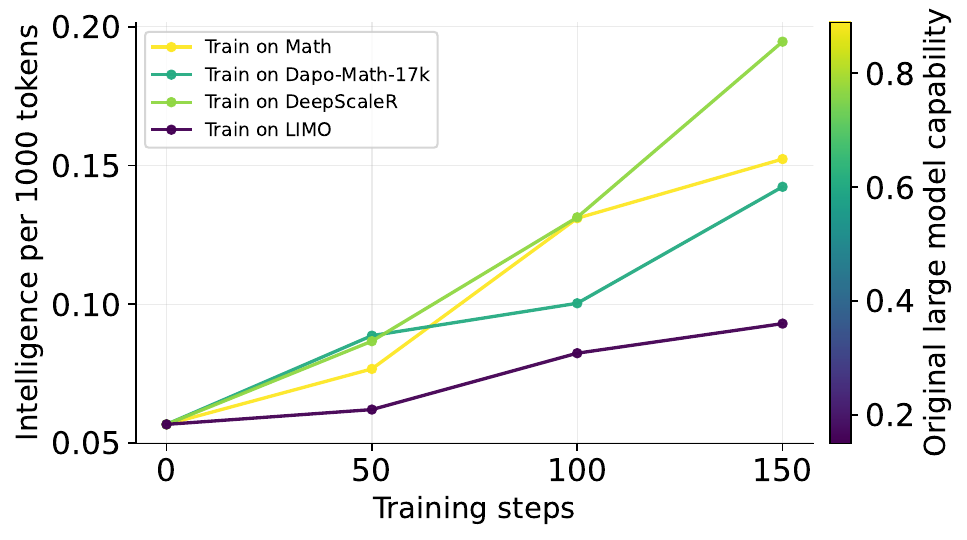}
    \caption{ \em The performance evolution from DUET  training. Each curve corresponds to a different training dataset, on which the large model exhibits a different level of accuracy. Lighter-colored curves indicate training datasets for which the large model has higher original accuracy. The y-axis reports the IPT metric averaged across all benchmarks except MATH500.}
    \label{fig:M_capability}
    \end{minipage}
    \hfill
    \begin{minipage}{0.49\textwidth}
    \centering
    \includegraphics[width=0.87\linewidth]{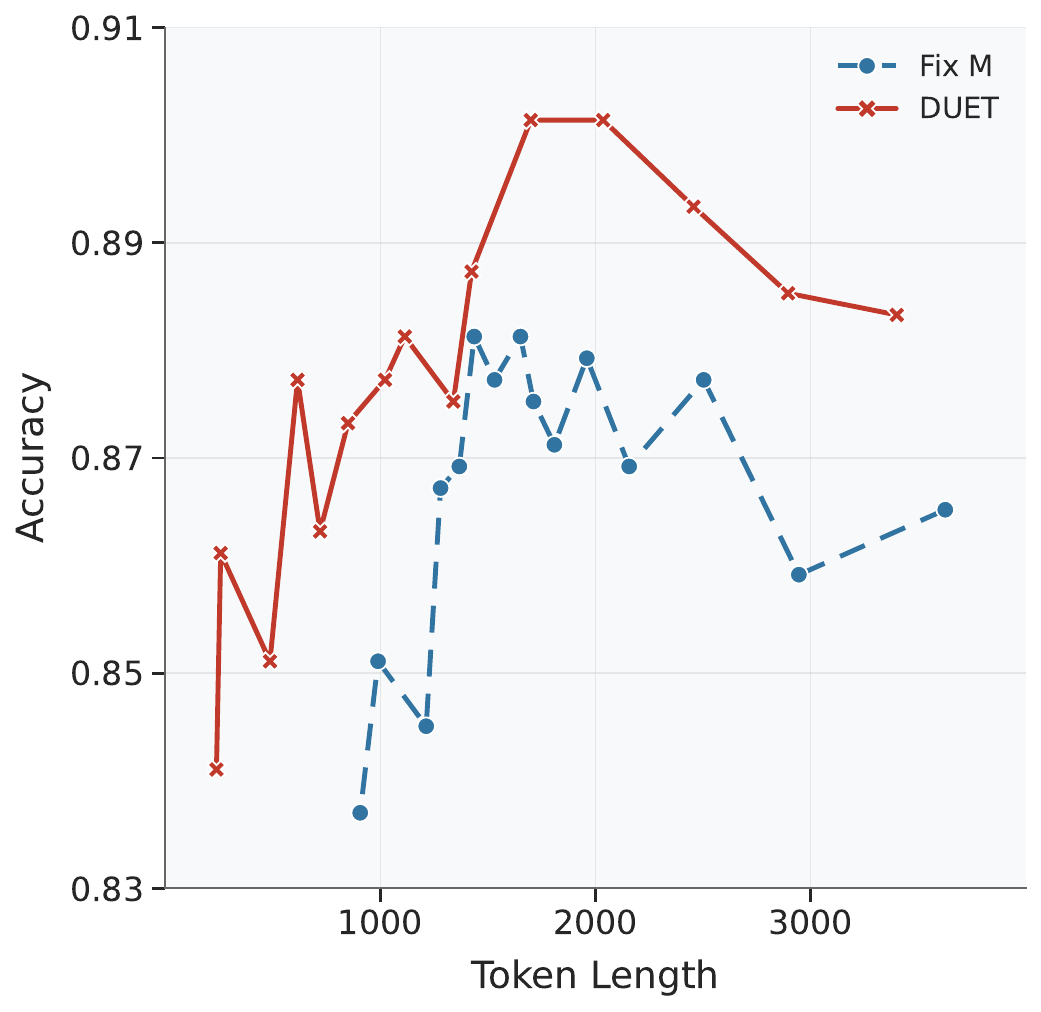}
    \caption{ \em Performance trajectories of \textit{DUET} and the \textit{Fix m} variant during training. Performance is logged every 10 training steps. The x-axis shows the total number of output tokens generated by the large model, and the y-axis reports task performance. Models are trained on DeepScaleR and evaluated on MATH500. Training proceeds from right to left.} 
    \label{fig:train_m}
    \end{minipage}
\end{figure}

To investigate this question, we trained DUET on the LIMO dataset~\citep{ye2025limoreasoning}, on which the original large model achieves only around $10–20\%$ accuracy. The results are shown in~\Cref{fig:M_capability}. When trained on LIMO, DUET fails to achieve meaningful compression. After 150 steps of training, DUET only improves IPT by a small margin compared to training on the other three datasets. The large model continues to produce long reasoning traces, and the overall token length remains high throughout training.

This behavior follows from the DUET objective in~\Cref{eq:model_obj}. When the large model lacks baseline capability on the training distribution, the marginal-utility signal $R(x,y_{Mm}) - R(x,y_m)$ becomes weak and noisy. Thus, without a reliable reasoning signal to compress, DUET cannot learn meaningful communication, leading the large model to either retain long traces or collapse toward uninformative outputs under the length penalty, as observed on the LIMO dataset.

\Cref{fig:M_capability} illustrates the evolution of the intelligence-per-token metric during training for DUET models trained on datasets where the large model exhibits varying levels of initial performance. Training on DeepScaleR yields the strongest gains, suggesting that the most effective training datasets are those on which the large model can solve problems confidently, yet which are not so easy as to limit opportunities for meaningful reasoning compression.

\subsubsection{Is it Necessary to Train the Small Model?}

In the default setting of DUET, the small model is trained jointly with the large model, allowing the two to co-evolve during optimization. To understand whether this co-training is essential, we also consider a variant in which the small model is kept fixed. In this setting, we keep the reward formulation unchanged but disable updates to the small model.

\Cref{fig:train_m} compares the evolution of the accuracy and token length on the MATH500 benchmark during training with a fixed small model (\textit{Fix $m$}) and with joint training (\textit{DUET}). Interestingly, we find that DUET remains effective even when the small model is not trained, still achieving meaningful compression of the large model’s reasoning. This suggests that part of DUET’s compression capability arises from optimizing the large model alone under the DUET objective. Further, throughout training, the DUET curve consistently lies above that of \textit{Fix $m$}, indicating a better trade-off between accuracy and inference efficiency. Jointly training the small model enables the large model to produce more concise reasoning without sacrificing performance. This behavior is intuitive: as the small model co-evolves with the large model, the large model can transmit increasingly compressed information that remains interpretable by the small model. In contrast, a fixed small model lacks the capacity to adapt to progressively more compact representations.

More importantly, training the small model makes our DUET framework applicable to broader scenarios. First, the role of the small model need not be limited to answer generation; it can be trained to serve other objectives or downstream tasks. Second, joint training becomes essential in settings that extend beyond the token space. For example, the large model may compress its reasoning into a latent representation, which the small model then decodes back into natural language. In such scenarios, training the small model is indispensable.

\subsubsection{Ablations on Marginal Utility and Length Penalty Schedule}
\label{subsec:ablation}
In~\Cref{sec:alg}, we introduce two practical design choices to stabilize the training: marginal utility and length penalty schedule. We conducted ablation experiments to study their effects, in which we removed each component individually or removed both. Specifically, without marginal utility means that the reward for the large model is $R(x,y_{Mm})$ alone without subtracting $R(x,y_{m})$ (see Section~\Cref{sec:alg} for details on this). Without the length penalty schedule means that a constant length-penalty coefficient is used throughout training.

\begin{wrapfigure}{r}{0.5\textwidth}
    \vspace{-7mm}
    \centering
    \includegraphics[width=1.\linewidth]{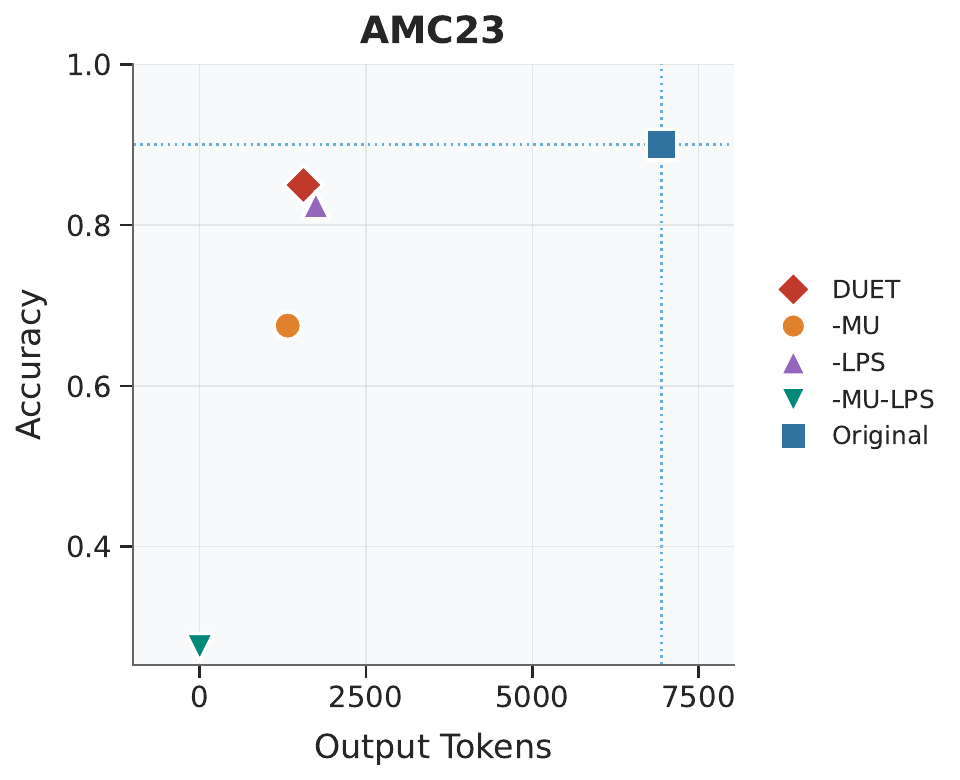}
    \caption{ \em The ablation results of marginal utility and length penalty schedule. –MU removes marginal-utility; –LPS removes the length-penalty schedule and sets a constant length-penalty coefficient as 0.5. Dashed blue vertical lines indicate the original large-model performance. The performance is evaluated on the AMC23 benchmark.}
    \label{fig:reward}
    \vspace{-7mm}
\end{wrapfigure}

\Cref{fig:reward} shows the ablation results. For each algorithmic variant, we evaluate the checkpoint after 150 training steps on the AMC23 benchmark. All models are trained on the Math-Lighteval dataset. The results show that, marginal utility is playing an important role while the length penalty schedule has a comparatively smaller impact, removing the schedule achieves similar performance to the default DUET configuration. 

When marginal utility is removed, the large model still learns to reduce its output length, however, it also incurs a noticeable drop in  accuracy which is not preferred. This observation is consistent with expectations. Math-Lighteval is a relatively easy training dataset, on which the small model has a decent accuracy (around 40\%) without any help from the large model. Hence, without marginal utility, the large model tends to output few tokens to avoid length penalty, while still achieving a moderate accuracy relying solely on the small model's ability. With marginal utility, this behavior of relying completely on the small model itself is not rewarded.

When both marginal utility and the length penalty schedule are omitted, the training collapses: the large model converges to producing empty or extremely short outputs. For the full ablation results, we refer readers to~\Cref{app:more_ablation}.

\section{Conclusions and Future Works}
In this work, 
we introduced \textsc{DUET} for efficient inference
by splitting it into two stages: a high-capacity model that outputs a compact reasoning signal, and a lightweight model that decodes it into the final response. We show that \textsc{DUET} can eliminate up to $60\%$ of large-model output tokens on AIME and GPQA without sacrificing accuracy. More broadly, our results shift the efficiency lens from compressing \emph{parameters} to compressing the \emph{information} a model must transmit. 

\paragraph{Limitation.} Due to limited computing resources, we evaluate DUET using Qwen3-4B as the large model and do not explore larger model scales. It would be interesting to see the behavior of DUET at larger model scales and under different model-size combinations; unfortunately, we do not have access to the computational resources required for this! In addition, we evaluate DUET only on tasks with outcome-based rewards. It would be interesting to extend the framework to settings involving supervised fine-tuning losses or process rewards. 

\paragraph{Future Work.} Looking ahead, \textsc{DUET} offers several natural extensions: replacing text-token communication with continuous codes (e.g., latent codes or soft prompts) \citep{hao2024training, zou2025latent} may increase compression and enable smoother end-to-end training, albeit with reduced interpretability. Moving from a one-shot hand-off to an interactive protocol where the small model queries the bigger model only under low confidence could further reduce computational overhead. More broadly, training a single large model to output a model-agnostic reasoning code that multiple heterogeneous small models can reliably decode, amortizes expensive inference across many lightweight, task-specific decoders, and forms an exciting avenue for future work. Finally, DUET is complementary to existing efficiency techniques and can be combined with methods such as Flash Attention, speculative decoding, etc.

\subsection*{Acknowledgments}
AS thanks Nived Rajaraman for useful discussions. SG acknowledges the support of the MathWorks Engineering Fellowship. YP and YC were supported in part by the NSF under grants DEB-2433726, ECCS-2317079, CCF-2200052, and IIS-1914792, by the DOE under grant DE-AC02-05CH11231, by the NIH under grant UL54 TR004130, and by the Hariri Institute at Boston University. The authors would also like to 
acknowledge New England Research Cloud (NERC) for providing the computational resources that
partially supported the experiments conducted in this work.

\newpage

\bibliographystyle{abbrvnat}
\bibliography{references} 

\newpage
\appendix

\onecolumn

\section{Additional Related Works} 

\label{app:related_work}
Here, we include additional related works that are not covered in the main paper.

\noindent \textbf{Model collaboration.} \asedit{A popular line of model collaboration research studies role-based collaboration, in which different models are assigned complementary responsibilities within a task.} \citet{juneja2023small} proposed a two-model system consisting of a decomposition generator that breaks a task into subproblems and a solver that addresses each subproblem. Similarly, \citet{grand2025self} introduced a planner--follower paradigm, where a planner model generates a task-specific inference program that is executed by a population of follower models. \citet{han2025uncertainty} proposed an uncertainty-aware collaborative system that offloads high-confidence inputs to a smaller model while reserving uncertain cases for a larger model. \asedit{Relatedly, \citet{zhang2025recursive} proposed Recursive Language Models, in which a model programmatically inspects long inputs and recursively invokes itself on selected subcontexts. While these methods emphasize decomposition and recursive control rather than explicit reasoning compression, they share with DUET the broader view that inference efficiency can arise from structuring computation across models or calls rather than relying on a single monolithic generation pass.}

\noindent \textbf{Knowledge distillation.} Knowledge distillation \citep{hinton2015distilling} is another form of model collaboration in which a large teacher model transfers knowledge to a smaller student model to improve efficiency. Several works \citep{ho2023large,hsieh2023distilling,magister2023teaching} have explored generating reasoning traces from powerful teacher models to fine-tune smaller students. On-policy distillation \citep{gu2023minillm,agarwal2024policy} also emerged as a popular method where \asedit{student models are} trained on data generated by themselves under the teacher's supervision.

\noindent \textbf{\asedit{Memory compaction and context compression.}} \asedit{Another related direction studies how to compress the model's working context or internal state rather than only its final answer tokens. \citet{zhang2025activationbeacon} proposed Activation Beacon, which compresses long contexts at the activation/KV-cache level to improve efficiency. More recently, \citet{monea2025breadcrumbs} introduced learned compression beacons that periodically summarize past KV states during reasoning, and \citet{wu2025resum} studied periodic context summarization for long-horizon agentic search. These approaches are complementary to DUET: they primarily compress memory or history within a single reasoning process, whereas DUET learns a concise reasoning signal that is explicitly optimized to be consumed by a second model (in an independent auto-regressive run).} 

\asedit{At the systems level, Cursor reports that Composer is trained to periodically self-summarize its working context to continue long-horizon problem solving under context constraints, while Meta reports that Muse Spark can launch multiple subagents in parallel for harder queries, illustrating a related trend toward system-level decomposition in deployed assistants \citep{cassano2026composer,meta2026musespark}.}

\noindent \textbf{Concise reasoning.} Towards concise reasoning, one popular line of work focuses on explicit budget control. Budget forcing methods terminate the reasoning process once a predefined token limit is reached \citep{muennighoff2025s1}. \citet{liu2025dler} studied the accuracy degradation caused by truncation, and showed that it arises from inadequate RL optimization rather than the lack of sophisticated penalties. Other approaches guide reasoning to adhere to a budget by predicting the remaining thinking length \citep{li2025steering} or dynamically adjusting the token budget based on problem complexity \citep{han2025token}. \asedit{\citet{gupta2026reasoncache} show that training a prefix cache while keeping the base model frozen is sufficient to improve reasoning performance, and strikingly induces more concise reasoning even without explicitly optimizing for brevity.}

\asedit{There are also several heuristic approaches.} These include prompt-based methods for eliciting concise reasoning \citep{nayab2024concise}, controlled frameworks that encourage step-skipping behavior to generate shorter yet accurate reasoning paths \citep{liu2024can}, and reinforcement learning paradigms that employ control tokens (e.g., \textless short\textgreater{} and \textless think\textgreater{}) to switch between concise and detailed reasoning modes \citep{fang2025thinkless}. \citet{lu2025retro} proposed Retro-Search, which retrospectively revises reasoning paths to discover higher-quality yet shorter traces, enabling student models to achieve faster inference with strong reasoning performance. \asedit{Relatedly, \citet{hou2025thinkprune} and \citet{xia2025tokenskip} explicitly prune or skip low-utility reasoning tokens to improve the performance--length trade-off.}

\noindent \textbf{\asedit{Latent reasoning and semantic communication.}} \asedit{A closely related line of work asks whether reasoning or inter-model communication should occur in natural language at all. \citet{hao2024continuous} introduced Chain of Continuous Thought, where intermediate reasoning unfolds in continuous latent space rather than through discrete text tokens. \citet{shen2025codi} compress explicit chain-of-thought into continuous representations via self-distillation, while \citet{su2025tokenassorted} mix latent and text tokens to shorten reasoning traces while preserving learnability. \citet{zhu2025superposition} provided theoretical support for such latent reasoning by showing that continuous thoughts can encode superpositions of multiple search frontiers. In multi-model settings, \citet{fu2025cachetocache}, \citet{zheng2025thoughtcommunication}, and \citet{zou2025latentmas} explore direct semantic communication through KV caches, latent thoughts, or shared latent working memory instead of textual messages. At a broader level, this perspective is closely connected to semantic communication, where the goal is to transmit task-relevant meaning rather than faithfully reproduce surface form \citep{wang2024llmsc}. Compared with these approaches, DUET deliberately retains a textual, inspectable communication channel between the large and small models, trading off some compression efficiency for interpretability and ease of training.}

\newpage
\section{Algorithm}
\label{app:algorithm}

\begin{algorithm}[htb]
\caption{DUET: Training Algorithm}  
\label{alg:duet_train}
\begin{algorithmic}[1]
\STATE {\bfseries Input:} Dataset $\mathcal{D}$, capable model $M_\theta$, lightweight model $m_\phi$, reward function $R(\cdot)$, initial length penalty weight $\lambda_0$, number of rollouts $G$, batch size $S$, training steps $T$, length penalty schedule factors $\eta$ and $B$. 
\FOR{$t=1$ {\bfseries to} $T$}
    \STATE Sample a minibatch $\{(x_i, y_i^*)\}_{i=1}^S \sim \mathcal{D}$ of $S$ samples, consisting of prompts $x$ and a response $y$ (containing the answer).  
    
    \FOR{$i=1$ {\bfseries to} $S$}
        \STATE Sample $G$ reasoning signals from the capable model: 
        \[
        z_{i,1:G} \sim M_\theta(\cdot \mid x_i).
        \]
        \STATE Sample a response from the lightweight model (without the reasoning signal): 
        \[
        y^m_{i} \sim m_\phi(\cdot \mid x_i).
        \]
        \STATE Compute baseline reward $r^m_{i} \leftarrow R(x_i, y^m_{i}, y_i^*)=\mathbf{1}\{y^m_i=y^*_i\}$. 
        \FOR{$j=1$ {\bfseries to} $G$}
            \STATE Generate a response from the lightweight model conditioned on $(x_i, z_{i,j})$:
            \[
            \hat y_{i,j} \sim m_\phi(\cdot \mid x_i, z_{i,j}).
            \]
            \STATE Compute reward $r_{i,j} \leftarrow R(x_i, \hat y_{i,j}, y_i^*)$ and reasoning length $\ell_{i,j} \leftarrow L(z_{i,j})$.
        \ENDFOR
    \ENDFOR
    
    \STATE Update the lightweight model $m_\phi$ using GRPO with the rewards: $r_{i,j}$.
    
    \STATE Update the capable model $M_\theta$ using GRPO with the rewards: $r_{i,j} - r^m_i - \lambda_{t-1} \,\ell_{i,j}/B$. 

    \STATE Update the length penalty coefficient $\lambda$:
    \[
    \lambda_{t} \leftarrow \lambda_{t-1} + \eta \left(\frac{E[L]}{B} - 1\right), \quad\text{and} \quad \  E[L]=\frac{1}{nS}\sum_{i=1}^S\sum_{j=1}^n \ell_{i,j}. 
    \]
\ENDFOR
\STATE {\bfseries Return:} Trained models $(M_\theta, m_\phi)$. 
\end{algorithmic}
\end{algorithm}
 
\newpage
\section{Experimental Setup}
\label{app:exp_setup}
All experiments are conducted on four NVIDIA H100 GPUs. We set hyperparameters $\eta=0.01$ and $B=1,000$. The training batch size is set to 64. The learning rate for both the large and small model is $1\times10^{-6}$. For both the large and small models, we sample four rollouts per input to compute advantages during GRPO training. We applied a KL-divergence regularization term to the reward with a coefficient of 0.001. The maximum response lengths for the large model $M$ and the small model $m$ are set to 16,384 and 2,048 tokens, respectively. The maximum prompt length is set to 1,024 tokens for the large model and 18,432 tokens for the small model. The rollout temperature for training and validation is set to 1.0 and 0.0, respectively. 

All training is implemented using VERL \citep{sheng2024hybridflow} as the underlying framework.

\section{Baseline Algorithms} 
\label{app:baseline}
We compare DUET against the following baseline algorithms: 
\begin{itemize}
 \item \textbf{Prompt-based:} In this baseline, we encourage concise reasoning through prompt engineering. Specifically, we augment the input with the instruction: “Note that you should keep the reasoning concise: include only essential reasoning steps and avoid repetition or unnecessary explanations.” 

\item \textbf{Single-model:} This baseline removes the small model and retains only the large model. The large model is trained using the same reward function as in DUET, but without the two-stage collaboration.

\item \textbf{ThinkPrune \citep{hou2025thinkprune}:} This method enforces a maximum response length during training. Only the portion of the response within the length limit is decoded to extract the final answer and compared against the ground truth. Consequently, a response receives a reward only if it produces the correct answer before exceeding the preset length. The maximum length is progressively reduced in a stepwise schedule: we initialize it at 8000 tokens and decrease it by 1000 every 20 training steps, with a minimum threshold of 1000 tokens.

\item \textbf{L-GRPO \citep{song2025walk}:} This baseline introduces a group-based length penalty. Specifically, a length penalty term is added to the reward, defined as the ratio between the response length and the maximum length among correct responses within the group. The penalty is applied only to correct responses.
\end{itemize}

\newpage
\section{Additional Experimental Details} 

\subsection{Total Cost} 

For a model with $B$ parameters, given a prompt of length $P$, the FLOPs required to generate a response of length $L$ can be approximated as 
$2BL + 4nd\left(LP + \frac{L(L-1)}{2}\right)$, where $n$ is the number of transformer layers and $d$ is the hidden dimension. In practice, the parameter-dependent term dominates, and the total cost can be well approximated by $2BL$. Therefore, throughout the paper, we use the output token length as a proxy for computational cost. We report only the large model’s output tokens in the main page, as its parameter size dominates the overall computation compared to the small model. For completeness, we provide the token counts of both models in Table~\ref{tb:output_tokens_both}.

In addition, we measure the wall-clock inference time on the full evaluation dataset. The original model requires 63m27.599s, while DUET reduces this to 41m54.337s, demonstrating a substantial improvement in practical efficiency.

\begin{table}[htb]
\centering
\caption{\em The performance and efficiency of DUET under different training steps. Each cell reports (accuracy, large-model tokens, small-model tokens).}
\label{tb:output_tokens_both}
\vspace{0.5em}
\begin{tabular}{lcccc}
\toprule
\multirow{2}{*}{Training steps}
& \multicolumn{4}{c}{Benchmark} \\
\cmidrule(lr){2-5}
& MATH 500 & AMC 23 & AIME 2024 & GPQA Diamond \\
\midrule
50
& (0.90, 1946, 402)
& (0.92, 3426, 538)
& (0.70, 7892, 705)
& (0.53, 4082, 6) \\
100
& (0.88, 1106, 452)
& (0.88, 1800, 634)
& (0.63, 5337, 837)
& (0.48, 2402, 13) \\
150
& (0.84, 753, 376)
& (0.80, 1504, 571)
& (0.53, 3681, 812)
& (0.48, 1292, 6) \\
\bottomrule
\end{tabular}
\end{table}

\subsection{Values Used to Plot Main Figures}
In this subsection, we report the numerical values used to generate the figures in the main paper, in order to facilitate reproducibility and detailed inspection of the results.

\begin{table}[htb]
\centering
\caption{ \em Numerical values used to plot Figure~\ref{fig:main}. Each cell reports (accuracy, large-model output tokens).}
\label{tb:number_fig_main}
\vspace{0.5em}
\begin{tabular}{lcccc}
\toprule
\multirow{3}{*}{Method}
& \multicolumn{4}{c}{Benchmark} \\
\cmidrule(lr){2-5}
& MATH & AMC & AIME & GPQA \\
& 500  & 23  & 2024 & Diamond \\
\midrule
\textbf{Original large model}
& (0.89, 4358)
& (0.90, 6948)
& (0.66, 12458)
& (0.53, 6928) \\
\midrule
\multicolumn{5}{l}{\textbf{Naive truncation with token budget:}} \\
\quad 3072
& (0.77, 3059)
& (0.65, 3792)
& (0.26, 4744)
& (0.32, 3097) \\
\quad 4096
& (0.83, 3445)
& (0.72, 4409)
& (0.36, 5561)
& (0.40, 3817) \\
\quad 8192
& (0.87, 4050)
& (0.87, 6266)
& (0.60, 8684)
& (0.50, 5900) \\
\midrule
\multicolumn{5}{l}{\textbf{DUET at training steps:}} \\
\quad 50
& (0.90, 1946)
& (0.92, 3426)
& (0.70, 7892)
& (0.53, 4082) \\
\quad 100
& (0.88, 1106)
& (0.88, 1800)
& (0.63, 5337)
& (0.48, 2402) \\
\quad 150
& (0.84, 753)
& (0.80, 1504)
& (0.53, 3681)
& (0.48, 1292) \\
\bottomrule
\end{tabular}
\end{table}

\begin{table}[htb]
\centering
\caption{ \em Numerical values used to plot Figure~\ref{fig:M_capability}. Each cell reports the intelligence-per-1000-tokens metric evaluated at 50, 100, and 150 training steps, respectively.}
\label{tb:number_fig_M}
\vspace{0.5em}
\begin{tabular}{lccc}
\toprule
\multirow{3}{*}{Training dataset}
& \multicolumn{3}{c}{Benchmark} \\
\cmidrule(lr){2-4}
& AIME & AIME & GPQA \\
& 2024  & 2025 & Diamond \\
\midrule
\textbf{MATH-LightEval}
& (0.075, 0.126, 0.121)
& (0.045, 0.073, 0.075)
& (0.113, 0.162, 0.230) \\
\midrule
\textbf{DeepScaleR}
& (0.089, 0.119, 0.109)
& (0.042, 0.079, 0.106)
& (0.129, 0.196, 0.369) \\
\midrule
\textbf{DAPO-MATH-17k}
& (0.085, 0.096, 0.124)
& (0.068, 0.043, 0.073)
& (0.113, 0.162, 0.230) \\
\midrule
\textbf{LIMO}
& (0.05, 0.074, 0.094)
& (0.046, 0.065, 0.055)
& (0.09, 0.108, 0.130) \\
\bottomrule
\end{tabular}
\end{table}

\begin{table}[htb]
\centering
\caption{ \em Numerical values used to plot Figure~\ref{fig:reward}. –MU removes marginal-utility; –LPS removes the length-penalty schedule. The performance is evaluated on the AMC23 benchmark.}
\label{tb:number_fig_reward}
\vspace{0.5em}
\begin{tabular}{lcc}
\toprule
Algorithm & Accuracy & Output tokens \\
\midrule
DUET & 0.85 & 1559 \\
\midrule
-MU & 0.67 & 1323 \\
\midrule
-LPS & 0.82 & 1746 \\
\midrule
-MU -LPS & 0.27 & 0 \\
\bottomrule 
\end{tabular}
\end{table}

\newpage
\subsection{Necessity of Training the Small Model}
In the main paper, we use Qwen3-0.6B as our small model. Despite having only 0.6 billion parameters, it is a relatively capable model that can generate accurate answers when conditioned on the large model's reasoning from the very beginning. We also experimented with Qwen2.5-0.5B-Instruct as the small model, which initially has limited ability to decode the large model's reasoning into correct answers. Through joint training, it quickly learns to solve problems by leveraging reasoning guidance from the large model, demonstrating that training the small model $m$ is necessary in this case. 

\begin{figure}[htb]
    \centering
    \includegraphics[width=0.75\linewidth]{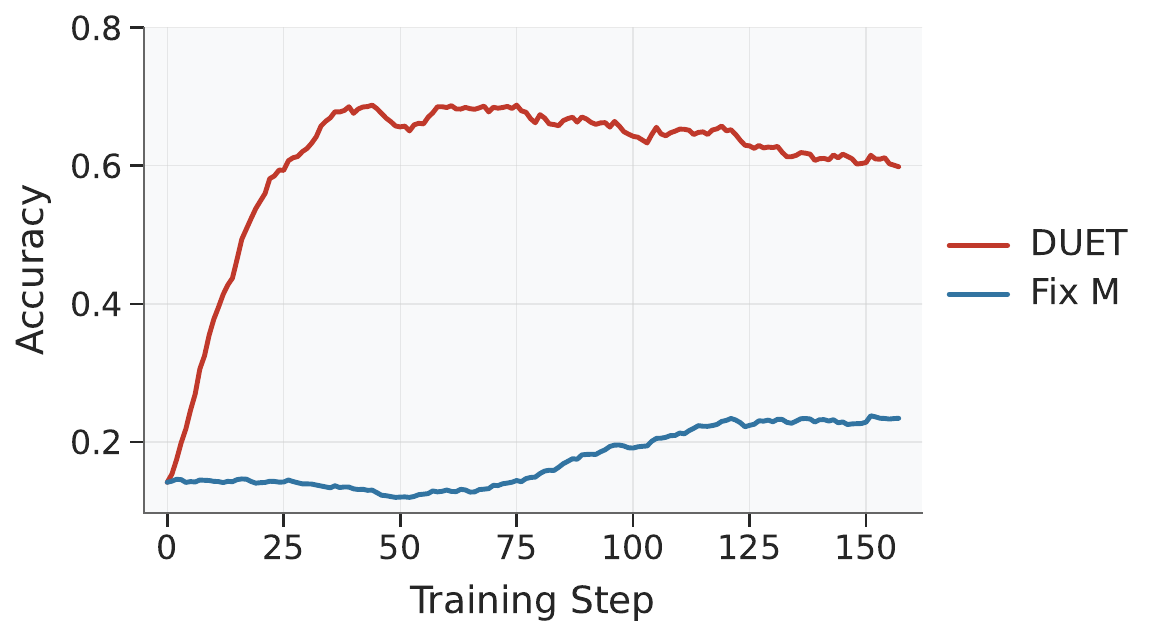}
    \caption{ \em Training dataset performance of DUET versus the fixed-$m$ variant during training. The small model is Qwen2.5-0.5B-Instruct and models are trained on the DeepScaleR dataset.}
    \label{fig:qwen05}
\end{figure}

Figure~\ref{fig:qwen05} shows the training progress of DUET compared to the variant with a fixed small model $m$. When the small model has limited initial capability, fixing $m$ causes it to struggle with generating correct responses based on the large model's reasoning, whereas DUET successfully trains the small model to generate correct answers through joint optimization.

Another interesting observation is that even without training the small model, it still shows slight performance improvement, suggesting that the large model adapts its reasoning to be more interpretable for the small model.

\subsection{Additional Ablation Results on Marginal Utility and Length Penalty Schedule}
\label{app:more_ablation}
\subsubsection{Trained on MATH}
\begin{figure*}[!htb]
    \centering

    \includegraphics[width=0.9\linewidth]{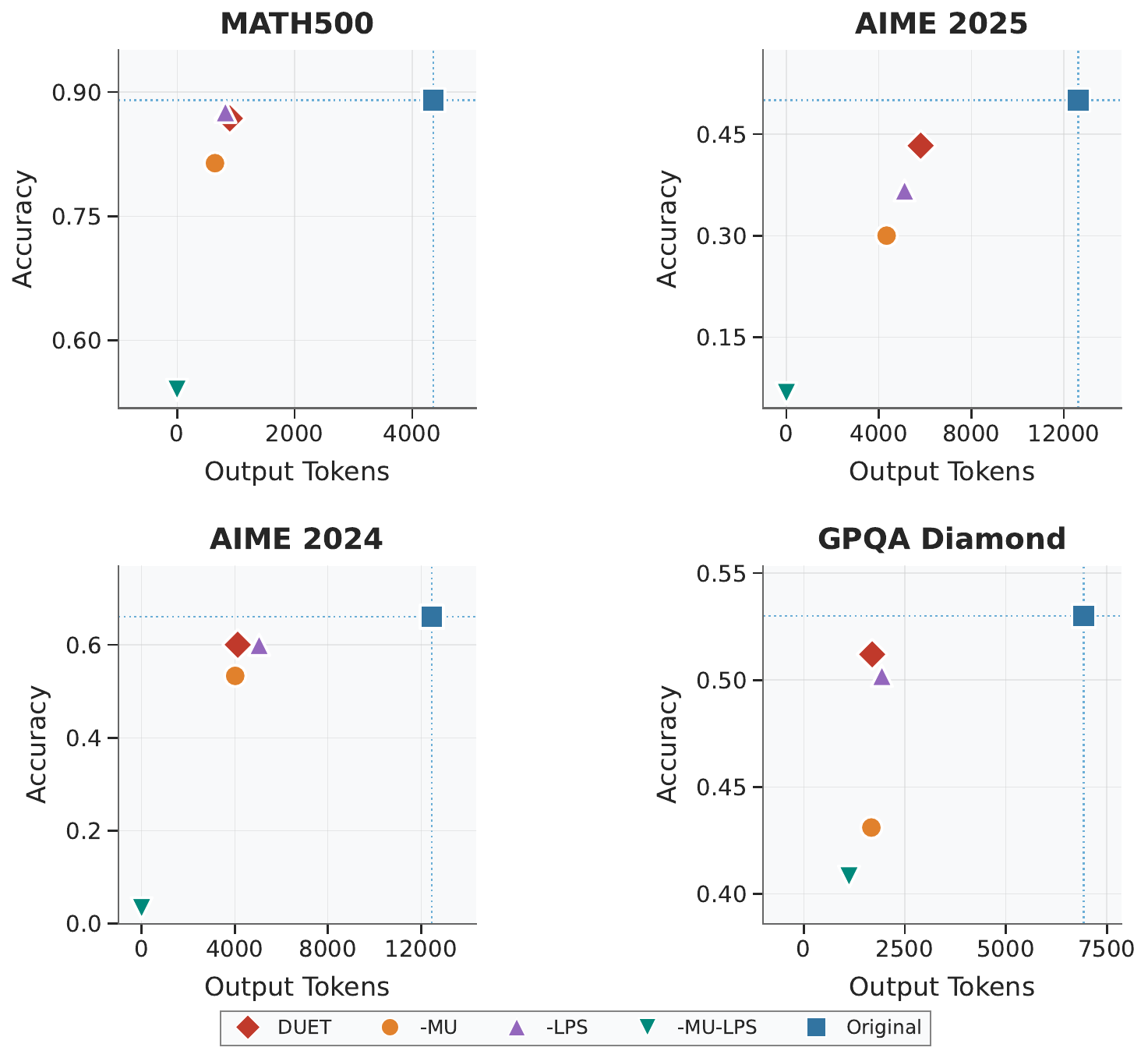}
    
    \caption{ \em The ablation results of marginal utility and length penalty schedule. –MU removes marginal-utility; –LPS removes the length-penalty schedule. Dashed blue vertical lines indicate the original large-model performance. The models are trained on MATH-LightEval dataset for 150 steps.}
    \label{fig:reward_ablation_math}
\end{figure*}

In Figure~\ref{fig:reward} in the main paper, we present ablation results evaluated on the AMC23 benchmark. Here in Figure~\ref{fig:reward_ablation_math}, we report the corresponding results on the remaining benchmarks. The observations are consistent with those in the main paper: marginal utility plays a more important role than the length penalty schedule. Using a constant length-penalty coefficient still yields reasonable performance, but the full DUET configuration achieves the best overall results. When both marginal utility and the length penalty schedule are removed, training becomes unstable and often collapses, causing the large model to converge to producing empty or near-empty outputs.

\subsubsection{Trained on DeepScaleR}

\begin{figure*}t[!htb]
    \centering

    \includegraphics[width=\textwidth]{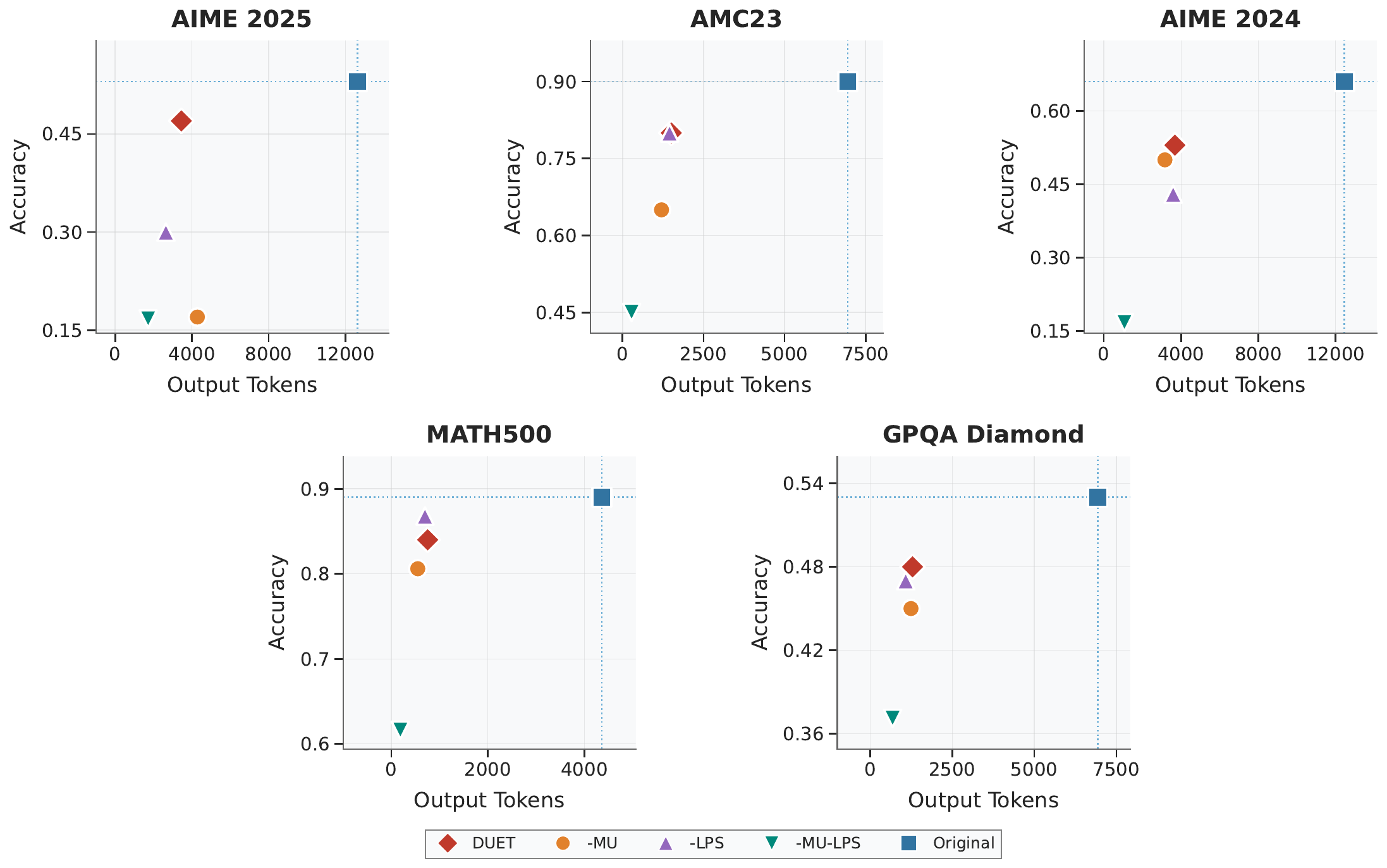}
    \caption{ \em The ablation results of marginal utility and length penalty schedule. –MU removes marginal-utility; –LPS removes the length-penalty schedule. Dashed blue vertical lines indicate the original large-model performance. The models are trained on DeepScaleR dataset for 150 steps.}
    \label{fig:reward_ablation_dsr}
\end{figure*}

Figures~\ref{fig:reward} and~\ref{fig:reward_ablation_math} present results for models trained on the MATH-LightEval dataset. In Figure~\ref{fig:reward_ablation_dsr}, we report the same ablation results when training on the DeepScaleR dataset. The overall observations are consistent across datasets. Marginal utility plays a more important role than the length penalty schedule—using a constant length-penalty coefficient still yields good performance on most benchmarks except AIME2025. And when both marginal utility and the length penalty schedule are removed, training becomes unstable and often collapses, causing the large model to converge to producing empty or near-empty outputs.

\subsection{Ablation on Training Dataset}

To verify that the observed behavior arises from the DUET framework and joint training rather than the dataset itself, we train the large model on the same training data using a standard reward objective with GRPO. The results are shown in Table~\ref{tb:data_ablation}. Training only the large model with a standard RL algorithm fails to produce concise reasoning, demonstrating that the ability to generate concise reasoning stems from the DUET framework and length-penalized joint training, rather than from the training dataset itself.

\begin{table}[htb]
\centering
\caption{ \em Large model performance across different training datasets using outcome reward only with GRPO. All models are evaluated at 100 training steps.}
\label{tb:data_ablation}
\vspace{0.5em}
\begin{tabular}{lccccccccccc}
\toprule
\multirow{2}{*}{Training dataset}
& \multicolumn{2}{c}{MATH500} 
& \multicolumn{2}{c}{AMC23}
& \multicolumn{2}{c}{AIME24} 
& \multicolumn{2}{c}{AIME25} 
& \multicolumn{2}{c}{GPQA Diamond} \\
\cmidrule(lr){2-3} \cmidrule(lr){4-5} \cmidrule(lr){6-7} \cmidrule(lr){8-9} \cmidrule(lr){10-11}
& Acc. & Tokens
& Acc. & Tokens
& Acc. & Tokens
& Acc. & Tokens
& Acc. & Tokens \\
\midrule
\textbf{Original large model}
& 89.4 & 4358
& 90.0 & 6948
& 66.7 & 12458 
& 53.3 & 12625
& 53.4 & 6928 \\
\midrule
\textbf{MATH-LightEval}
& 90.4 & 4702
& 90.0 & 7410
& 66.7 & 11233 
& 60.0 & 13125 
& 59.9 & 7345 \\
\midrule
\textbf{DeepScaleR}
& 89.0 & 4739
& 97.5 & 7108
& 73.3 & 12750
& 60.0 & 13635 
& 52.8 & 7163 \\
\midrule
\textbf{DAPO-MATH-17k}
& 89.6 & 4789
& 90.0 & 7569
& 60.0 & 11716
& 60.0 & 12857 
& 51.8 & 7186 \\
\bottomrule
\end{tabular}
\end{table}

\subsection{Ablation on Hyperparameters}
We conduct ablation studies on the hyperparameters in the length penalty, including $B$ and $\lambda$ (with constant schedule).

\subsubsection{The Value of $B$}
The hyperparameter $B$ is the length normalization factor which controls the strength of length penalty. We evaluate three values: $B \in \{500, 1000, 2000\}$, where $B=1000$ is the default used in the main paper. Results are shown in Table~\ref{tb:ablation_B_deepscaler} (trained on MATH-LightEval) and Table~\ref{tb:ablation_B_math} (trained on DeepScaleR). Smaller values of $B$ apply stronger length penalties. When $B=2000$, task performance is slightly higher than $B=500$ or $B=1000$, but produces more tokens. Conversely, when $B=500$, reasoning is most compressed but sacrifices more performance. Overall, DUET demonstrates stable performance across all tested values of $B$ in the range $[500, 2000]$.

\begin{table}[htb]
\centering
\caption{ \em Ablation study of hyperparameter $B$. Models are trained on the \textbf{DeepScaleR} dataset and evaluated at 100 training steps across multiple benchmarks. Acc. denotes accuracy (\%) and Tokens denotes average output tokens per sample.}
\label{tb:ablation_B_deepscaler}
\vspace{0.5em}
\begin{tabular}{lccccccccccc}
\toprule
\multirow{2}{*}{Value of B}
& \multicolumn{2}{c}{MATH500} 
& \multicolumn{2}{c}{AMC23}
& \multicolumn{2}{c}{AIME24} 
& \multicolumn{2}{c}{AIME25} 
& \multicolumn{2}{c}{GPQA Diamond} \\
\cmidrule(lr){2-3} \cmidrule(lr){4-5} \cmidrule(lr){6-7} \cmidrule(lr){8-9} \cmidrule(lr){10-11}
& Acc. & Tokens
& Acc. & Tokens
& Acc. & Tokens
& Acc. & Tokens
& Acc. & Tokens \\
\midrule
\textbf{$B=500$}
&87.0  &954 
&77.5  &2054 
&50.0  &4980  
&36.7  &4876 
&45.2  &1914 \\
\midrule
\textbf{$B=1000$}
&88.0  &1106
&87.5  &1801
&63.3  &5338 
&46.7  &5888
&47.2  &2403 \\
\midrule
\textbf{$B=2000$}
&88.8  &1375
&90.0  &2256
&53.3  &6095 
&46.7  &6624
&51.3  &2644 \\
\bottomrule
\end{tabular}
\end{table}

\begin{table}[htb]
\centering
\caption{ \em Ablation study of hyperparameter $B$. Models are trained on the \textbf{MATH} dataset and evaluated at 100 training steps across multiple benchmarks. Acc. denotes accuracy (\%) and Tokens denotes average output tokens per sample.}
\label{tb:ablation_B_math}
\vspace{0.5em}
\begin{tabular}{lccccccccccc}
\toprule
\multirow{2}{*}{Value of B}
& \multicolumn{2}{c}{MATH500} 
& \multicolumn{2}{c}{AMC23}
& \multicolumn{2}{c}{AIME24} 
& \multicolumn{2}{c}{AIME25} 
& \multicolumn{2}{c}{GPQA Diamond} \\
\cmidrule(lr){2-3} \cmidrule(lr){4-5} \cmidrule(lr){6-7} \cmidrule(lr){8-9} \cmidrule(lr){10-11}
& Acc. & Tokens
& Acc. & Tokens
& Acc. & Tokens
& Acc. & Tokens
& Acc. & Tokens \\
\midrule
\textbf{$B=500$}
&86.4  &776 
&85.0  &1906 
&56.7  &3472  
&33.3  &5918 
&47.7  &2221 \\
\midrule
\textbf{$B=1000$}
&89.8  &1110
&85.0  &1879
&63.3  &5014 
&46.7  &6419
&52.3  &2690 \\
\midrule
\textbf{$B=2000$}
&91.0  &2019
&90.0  &3923
&70.0  &7424 
&46.7  &9438
&50.3  &4894 \\
\bottomrule
\end{tabular}
\end{table}

\subsubsection{Length Penalty Coefficient}
In the main paper, we adopt an adaptive length penalty schedule. Here, we also compare it with constant penalty schedules using different coefficients. The results are summarized in Table~\ref{tb:ablation_lambda_math}. Larger values of $\lambda$ apply stronger length penalties, achieving more compressed reasoning. As shown in the ablations on marginal utility and length penalty schedule (Section~\ref{app:more_ablation}), the constant schedule achieves similar performance to the adaptive schedule, demonstrating DUET's robustness to the choice of length penalty coefficient. Furthermore, performance remains stable across different values of $\lambda$ in the range $[0.1, 2.0]$, which further confirms the stability of DUET.

\begin{table}[htb]
\centering
\caption{ \em Ablation study of hyperparameter $\lambda$ when using \textbf{constant length penalty schedule}. Models are trained on the \textbf{MATH} dataset and evaluated at 100 training steps across multiple benchmarks. Acc. denotes accuracy (\%) and Tokens denotes average output tokens per sample.}
\label{tb:ablation_lambda_math}
\vspace{0.5em}
\begin{tabular}{lccccccccccc}
\toprule
\multirow{2}{*}{Value of $\lambda$}
& \multicolumn{2}{c}{MATH500} 
& \multicolumn{2}{c}{AMC23}
& \multicolumn{2}{c}{AIME24} 
& \multicolumn{2}{c}{AIME25} 
& \multicolumn{2}{c}{GPQA Diamond} \\
\cmidrule(lr){2-3} \cmidrule(lr){4-5} \cmidrule(lr){6-7} \cmidrule(lr){8-9} \cmidrule(lr){10-11}
& Acc. & Tokens
& Acc. & Tokens
& Acc. & Tokens
& Acc. & Tokens
& Acc. & Tokens \\
\midrule
Adaptive schedule
&89.8  &1110
&85.0  &1879
&63.3  &5014 
&46.7  &6419
&52.3  &2690 \\
\midrule
Constant schedule: \\
\qquad \textbf{$\lambda=0.1$}
&90.4  &2278 
&92.5  &4063 
&66.7  &9302  
&66.7  &9500 
&48.7  &5002 \\
\qquad
\textbf{$\lambda=0.5$}
&90.6  &1267
&87.5  &2623
&60.0  &5862 
&43.3  &7304
&48.7  &2839 \\
\qquad \textbf{$\lambda=1.0$}
&90.2  &1005
&82.5  &1768
&53.3  &4923 
&40.0  &6584
&47.7  &2506 \\
\qquad
\textbf{$\lambda=2.0$}
&87.0  &871
&87.5  &1388
&56.7  &4713 
&53.3  &5552
&46.2  &2000 \\
\bottomrule
\end{tabular}
\end{table}

\subsection{The Influence of KL Penalty in the Reward}
In training, we include a KL penalty term in the reward for the large model, which is standard practice in the RL post-training paradigm. The KL penalty is computed as $\text{KL}(M_t \| M_0)$, where $t$ denotes the training step and $M_0$ is the initial model. We conduct ablations on the effect of this KL term, with results shown in Tables~\ref{tb:kl} and~\ref{tb:kl_nomulps_math}.  We find that including the KL penalty achieves better performance and training stability in DUET, while without the KL penalty, it still achieves similar performance. However, when removing marginal utility and switching from the adaptive length penalty schedule to a constant schedule, including the KL penalty can prevent training collapse to some extent, though performance remains significantly worse than DUET with default settings.

\begin{table}[htb]
\centering
\caption{ \em Effect of KL penalty on DUET performance. Models are trained on the MATH-LightEval dataset and evaluated at 100 training steps. Acc. denotes accuracy (\%) and Tokens denotes average output tokens per sample.}
\label{tb:kl}
\vspace{0.5em}
\begin{tabular}{lccccccccccc}
\toprule
\multirow{2}{*}{}
& \multicolumn{2}{c}{MATH500} 
& \multicolumn{2}{c}{AMC23}
& \multicolumn{2}{c}{AIME24} 
& \multicolumn{2}{c}{AIME25} 
& \multicolumn{2}{c}{GPQA Diamond} \\
\cmidrule(lr){2-3} \cmidrule(lr){4-5} \cmidrule(lr){6-7} \cmidrule(lr){8-9} \cmidrule(lr){10-11}
& Acc. & Tokens
& Acc. & Tokens
& Acc. & Tokens
& Acc. & Tokens
& Acc. & Tokens \\
\midrule
w/ KL penalty
&89.8  &1110
&85.0  &1879
&63.3  &5014 
&46.7  &6419
&52.3  &2690  \\
\midrule
w/o KL penalty
&89.2  &1174
&90.0  &2404
&53.3  &6084 
&43.3  &7295
&51.8  &2624 \\
\bottomrule
\end{tabular}
\end{table}

\begin{table}[htb]
\centering
\caption{ \em Effect of KL penalty on the ablated variant without marginal utility and adaptive length penalty schedule. Models are trained on the MATH-LightEval dataset and evaluated at 150 training steps. Acc. denotes accuracy (\%) and Tokens denotes average output tokens per sample.}
\label{tb:kl_nomulps_math}
\vspace{0.5em}
\begin{tabular}{lccccccccccc}
\toprule
\multirow{2}{*}{}
& \multicolumn{2}{c}{MATH500} 
& \multicolumn{2}{c}{AMC23}
& \multicolumn{2}{c}{AIME24} 
& \multicolumn{2}{c}{AIME25} 
& \multicolumn{2}{c}{GPQA Diamond} \\
\cmidrule(lr){2-3} \cmidrule(lr){4-5} \cmidrule(lr){6-7} \cmidrule(lr){8-9} \cmidrule(lr){10-11}
& Acc. & Tokens
& Acc. & Tokens
& Acc. & Tokens
& Acc. & Tokens
& Acc. & Tokens \\
\midrule
w/ KL penalty
&84.2  &684
&77.5  &1394
&40.0  &3945 
&30.0  &4116
&41.1  &1688  \\
\midrule
w/o KL penalty
&54.0  &0
&27.5  &0
&3.3  &0 
&6.7  &0
&47.7  &1121 \\
\bottomrule
\end{tabular}
\end{table}

\newpage
\subsection{Effect of Clipping the Length Penalty Coefficient}
\begin{figure}[htb]
    \centering
    \includegraphics[width=0.9\linewidth]{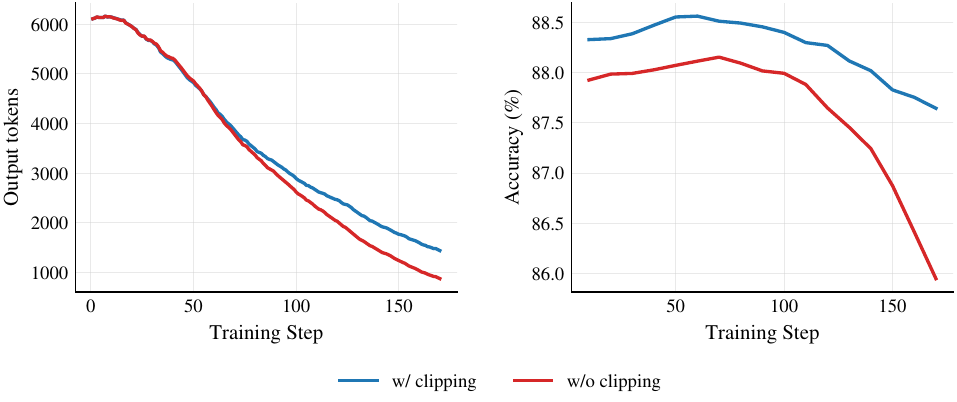}
    \caption{ \em Comparison of training with and without $\lambda$ clipping. Models are trained on the DeepScaleR dataset and evaluated on MATH500 benchmark.}
    \label{fig:clipping}
\end{figure}

In our algorithm, we apply a clipping mechanism on $\lambda$ to prevent explosive length penalties:
$$\lambda_{t} \leftarrow \max\{0, \min\{\lambda_t, 1\}\}.$$
Here, we explore the effect of this clipping mechanism. Figure~\ref{fig:clipping} shows task performance and output tokens throughout training. In the early training stage, clipping has minimal effect since $\lambda$ has not yet grown to large values. In the later training stage, clipping helps prevent training collapse and maintains good performance compared to the unclipped variant.

\end{document}